\documentclass[letterpaper]{article} 

\usepackage[submission]{aaai23}  
\usepackage{times}  
\usepackage{helvet}  
\usepackage{courier}  
\usepackage[hyphens]{url}  
\usepackage{graphicx} 
\usepackage{natbib}  
\usepackage{caption} 
\usepackage{algorithm}
\usepackage{algorithmic}
\usepackage{dsfont}
\usepackage{makecell}
\usepackage{multirow}
\usepackage{subfigure}
\usepackage{amsmath}
\usepackage{color}
\usepackage{amssymb}
\usepackage{paralist}
\usepackage{booktabs}
\usepackage{newfloat}
\usepackage{listings}
\DeclareCaptionStyle{ruled}{labelfont=normalfont,labelsep=colon,strut=off} 
\floatstyle{ruled}
\newfloat{listing}{tb}{lst}{}
\floatname{listing}{Listing}
\def\boxit#1{\vbox{\hrule\hbox{\vrule\kern6pt
			\vbox{\kern6pt#1\kern6pt}\kern6pt\vrule}\hrule}}

\title{TCFimt: Temporal Counterfactual Forecasting from Individual Multiple Treatment Perspective}
\author {
    First Author Name,\textsuperscript{\rm 1}
    Second Author Name, \textsuperscript{\rm 2}
    Third Author Name \textsuperscript{\rm 1}
}
\affiliations {
    \textsuperscript{\rm 1} Affiliation 1\\
    \textsuperscript{\rm 2} Affiliation 2\\
    firstAuthor@affiliation1.com, secondAuthor@affilation2.com, thirdAuthor@affiliation1.com
}

\usepackage{bibentry}

\begin{document}

\maketitle
\begin{abstract}
Determining causal effects of temporal multi-intervention assists decision-making. Restricted by time-varying bias, selection bias, and interactions of multiple interventions, the disentanglement and estimation of multiple treatment effects from individual temporal data is still rare. To tackle these challenges, we propose a comprehensive framework of temporal counterfactual forecasting from an individual multiple treatment perspective (TCFimt). TCFimt constructs adversarial tasks in a seq2seq framework to alleviate selection and time-varying bias and designs a contrastive learning-based block to decouple a mixed treatment effect into separated main treatment effects and causal interactions which further improves estimation accuracy. Through implementing experiments on two real-world datasets from distinct fields, the proposed method shows satisfactory performance in predicting future outcomes with specific treatments and in choosing optimal treatment type and timing than state-of-the-art methods.
  
\end{abstract}

\section{Introduction}

\begin{figure}
    \centering
    \includegraphics[scale=0.23]{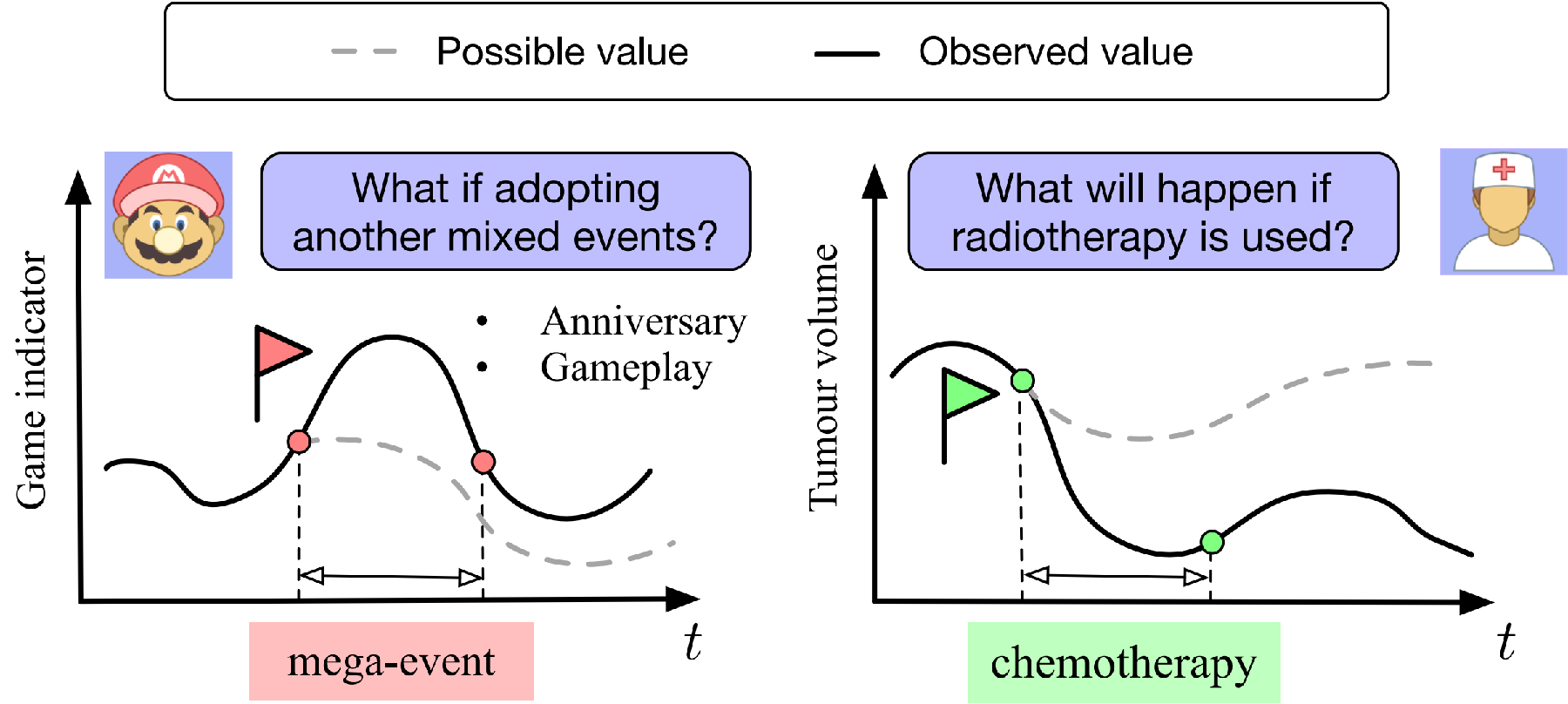}
    \caption{Counterfactual forecasting in real-world cases.}
    \label{fig:toy_example}
\end{figure}

Causal analysis in temporal scenarios is to explore which factors lead to the results in the future, which could assist in decision-making. In real-life scenarios, decision-makers are often faced with the dilemma of which decision or even a combination of options to choose, hence it is particularly important for them to reliably estimate the effects of distinct interventions and intervention interactions. For example, in the game field, game companies are really concerned about the changing revenue, which reflects various states of a game. As shown on the left of Figure~\ref{fig:toy_example}, when a noticeable change in a game indicator is observed, which is due to several mixed mega-events (e.g., game anniversary, the newly opened game server, and new gameplay) going on during this period. To weigh costs and benefits, they are eager to know what will happen if adopting other mixed interventions. Similarly, in the field of health care, doctors pay more attention to changes in vital signs, leading to what treatment regimens are needed. However, due to the physical characteristics of different patients, the treatment plans should be varied. The doctors also want to know what will happen in if the treatment is changed (right panel of Figure~\ref{fig:toy_example}). The above questions motivate our research on counterfactual forecasting of mixed interventions on time-series data in this paper.


   

Existing causal inference techniques under the longitudinal setting such as Recurrent Marginal Structural Networks \cite[RMSN,][]{lim2018forecasting} and Counterfactual Recurrent Network~\cite[CRN,][]{bica2020estimating} cannot be applied to conduct causal inference of multiple treatments in temporal data. First of all, only one of the multiple treatment result (factual outcome) can be observed, while the results of other interventions at that moment (counterfactual outcomes) are not available. This makes us never obtain the entire vector of outcomes. Second, the observational data is prone to multiple treatment selection biases. For example, it is easier for the rich to get better treatment, which causes the distribution of features among patients to vary drastically across different choices of treatments. Third, these observed outputs are the result of mixed effects of multiple interventions, making it difficult to estimate the effects of a single intervention and the causal interactions. 



Accordingly, to solve the above-mentioned challenges, this paper designs a Temporal Counterfactual Forecasting network from an Individual Multiple Treatment perspective (TCFimt), for enhancing prediction accuracy and inference effectiveness. Specifically, we consider the following settings to address these issues: 1) For unobserved counterfactual data, we design a corruption function to generate the pseudo-counterfactual data of multiple treatments. 2) Moreover, by simultaneously maximizing the error of the treatment classification task and minimizing the error of the outcomes prediction task, the selection bias and time-varying bias are alleviated by employing an adversarial training way. 3) Finally, an effect disentanglement block based on contrastive learning is dedicated to further enhancing the prediction of effects under mixed interventions.
The contributions of this research are summarized as follows:



\begin{compactitem}
    \item To our best knowledge, our paper is the first to explore the interventional effect decoupling and causal interaction issue in temporal data, which is an attempt at causal inference in complex scenarios.
    \item The proposed TCFimt method combines the way of adversarial training and contrastive learning, which is beneficial to solving the issues of temporal counterfactual forecasting with mixed interventions in one treatment plan.
    \item We provide a theoretical analysis of algorithms for learning the balanced representation in temporal individual multiple treatment effects estimation.
    \item The implemented experiments on real-world data in two different fields validate the effectiveness of our methods.
\end{compactitem}

\section{Preliminary}

In this section, we will employ the following notations\footnote{
Complete background on treatment effect inference is provided in supplementary material due to the length limit.
}. For time any $t>0$, we denote time-varying feature variables by $\mathbf{X}_t\in \mathcal{X}\subset \mathbb{R}^{D_x}$, a binary time-dependent treatment action by $\mathbf{A}_t=\{A_{1,t},\ldots,A_{K,t}\} \in \{0,1\}^{K}$, where, for $k=1,\ldots,K$, $A_{k,t}=1$ if treatment $k$ is received and $0$ if not at time $t$, and the corresponding interventional features by $\mathbf{V}_t = (V_{1,t},\ldots,V_{K,t})\in \mathcal{V}\subset \mathbb{R}^{D_v \times K}$, where $V_{k}$ denotes the features of treatment $k$. In particular, $K>2$ means the outcome is caused by multiple interventions. It is worth mentioning that, instead of using one-hot treatments of the mixed interventions to correspond to one intervention plan, we explore each of the intervention actions that make up the mixed intervention regimen. 

Following the potential outcomes framework proposed by Rubin in 1978~\cite{rubin1978bayesian} and extended in 2008~\cite{robins2008estimation} to account for time-varying treatments, we let $Y[\mathbf{a}_{t}]$ denote the individual potential outcome at time $t$ for the treatment $\mathbf{A}_t=\mathbf{a}_t$, and $Y_t = Y[\mathbf{A}_t]$ is the individual observed outcome at time $t$ of treatment $\mathbf{A}_t$. Consider a random sample of size $N$, for each individual $i=1,\ldots,N$, we observe discrete time series data $\{\mathbf{V}_s^{(i)}, \mathbf{A}_s^{(i)}, \mathbf{X}_s^{(i)}, Y_s^{(i)}\}_{s=1}^t$ for some positive integer $t$, distributed as $\{\mathbf{V}_s, \mathbf{A}_s, \mathbf{X}_s, Y_s\}_{s=1}^t$. 

To adapt the unconfoundedness assumption, we let $\overleftarrow{\mathbf{H}}_{t} = (\overleftarrow{A}_{t-1}, \overleftarrow{V}_{t}, \overleftarrow{X}_{t})$, where $\overleftarrow{A}_{t-1}=(\mathbf{A}_1,\ldots,\mathbf{A}_{t-1})$ is the historical treatment assignments, $\overleftarrow{V}_{t} = (V_1, ..., V_{t})$ is the adopted interventional feature up to time $t$, and $\overleftarrow{X} = (X_1, ..., X_t)$ is the time-varying state vector. 

\textbf{Assumption 1. (unconfoundedness)} Base on given notations, unconfoundedness assumption at time $t$ is defined as:
\begin{equation}
\{Y[\mathbf{a}_{t}]\}_{\mathbf{a}_t\in\{0,1\}^K} \perp \!\!\! \perp A_{t} \mid \overleftarrow{\mathbf{H}}_{t} \,. 
\end{equation}
Our goal is to estimate the conditional average treatment effect (CATE) for each treatment $k$, $k=1,\ldots,K$, in the future. For notational simplicity in defining our CATE, we let $Y[a_{k,t}]$ be the potential outcome $Y[\mathbf{a}_t]$ with treatment $k$ being $a_{k,t}\in\{0,1\}$ and $a_{j,t}=0$ for all $j\neq k$, and let $Y[a_{0,t}]$ be the potential outcome with none of the $K$ treatments (i.e. $a_{1,t}=\ldots=a_{K,t}=0$). Then, the CATE we aim to estimate is defined as
\begin{small}
\begin{equation}
\delta[a_{k,t+\tau}] =\mathbb{E}(Y[a_{k,t+\tau}] \mid \overleftarrow{\mathbf{H}}_{t})-\mathbb{E}(Y[a_{0,t+\tau}] \mid \overleftarrow{\mathbf{H}}_{t})\,,
\end{equation}
\end{small}
for $a_{k,t+\tau}=1$, $k=1,\ldots,K$, $\tau\geq 0$ an integer, where $t+\tau$ represents a future time step. We hope the future individual potential outcome could be estimated when we decide on specific treatment $a_{k,t+\tau}$, to facilitate decision making. For this purpose, we introduce the following assumption.

\textbf{Assumption 2. (causal interactions)} Assuming that the potential outcomes of multiple treatments can be divided into separated treatment effects and causal interactions. That is, we can define the causal interaction for any combination of the $K$ treatments $\mathbf{a}_t\in\{0,1\}^K$, $\delta_{CI}[\mathbf{a}_t]$, as follows:
\begin{small}
\begin{equation}
\begin{aligned}
\delta_{CI}[\mathbf{a}_t] &=\mathbb{E}((Y[\mathbf{a}_t] - Y[a_{0 ,t}])\mid \overleftarrow{\mathbf{H}}_{t})\\
&-\sum^K_{k=1} \mathbb{E}((Y[a_{k,t}]- Y[a_{0,t}]) \mid \overleftarrow{\mathbf{H}}_{t})\,.
\end{aligned}
\end{equation}
\end{small}

\section{Method}

\UseRawInputEncoding

Most existing methods estimate individual treatment effects in static setting without considering the time-varying confounders. Those temporal causal inference methods focus on single treatment regimen, ignoring realistic situations of multiple treatment assignments. More importantly, the causal interactions under mixed treatments is difficult to be accurately estimated. Considering the challenges of time-varying bias, multiple treatment selection  bias, and causal interactions in time series prediction task, in this section, we propose TCFimt to forecast the future target value under mixed treatments and quantify the individual treatment effects for decision making.

\begin{figure}
    \centering
    \includegraphics[scale=0.5, height = 3in]{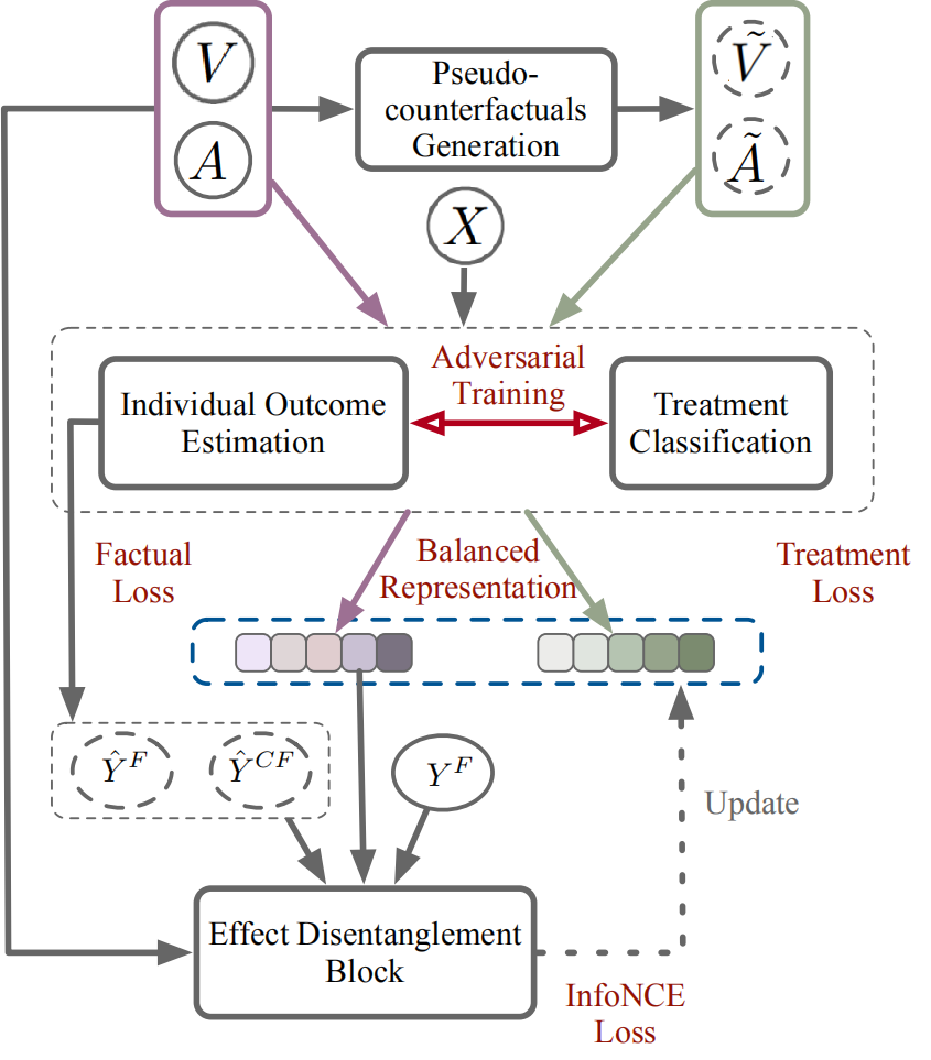}
    \caption{The architecture of our proposed method TCFimt.}
    \label{fig:whole_framework}
\end{figure}

The whole framework could be divided into three main parts, as shown in Figure~\ref{fig:whole_framework}. To solve the unobserved counterfactual data problem, we first design a corruption function to generate pseudo-counterfactual data with treatment actions and interventional features. The corruption function design also plays an auxiliary role in subsequent intervention effects separation. Second, we solve the time-varying bias issue and selection bias issue by conducting domain adversarial training in the individual potential outcome estimation task and the multiple treatment action classification tasks, then the balanced representation which is invariant to the treatment at present moment can be learned. In particular, individual outcomes are estimated through an encoder-decoder framework, whose decoder module supports the multi-horizon prediction. Note that due to the similar inputs and shared parameters, only employing the individual outcome estimate module cannot well decouple the colliding intervention effects. We propose to perform balanced representation learning and prediction on both factual and pseudo-counterfactual data. By applying the contrastive learning of factual and pseudo-counterfactual outcome and interaction effect assumption, our method decouple the mixed treatment effect effectively. In the later section, we theoretically show that such a strategy improves estimation accuracy under the setting of multiple treatments. In the third part, an effect disentanglement block is applied. By modeling the relationship between treatments and corresponding potential outcomes via the contrastive learning strategy, this block further highlights the effects of current interventions in the training process.

\begin{figure*}
    \centering
    \includegraphics[width=0.7\linewidth]{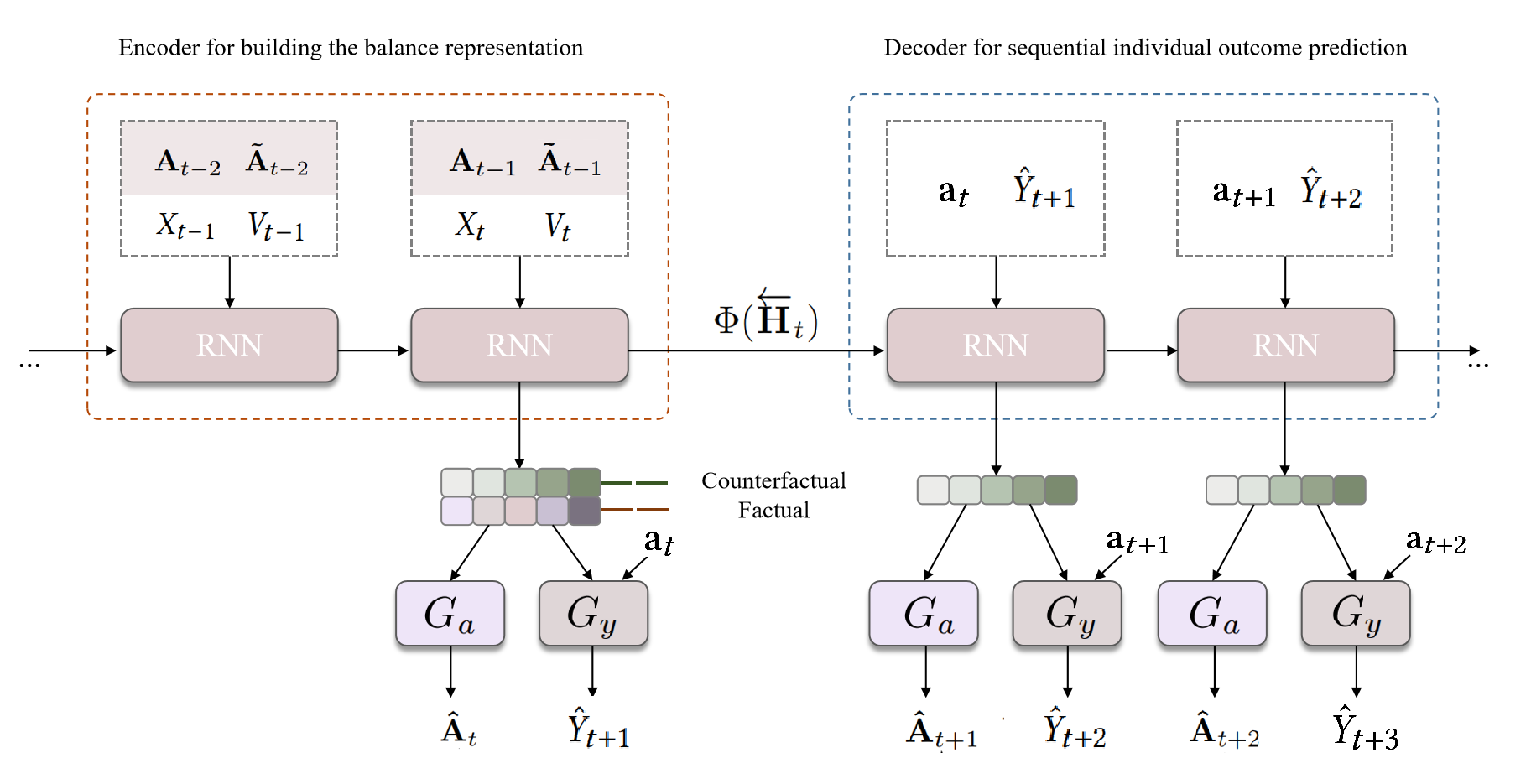}
    \caption{Encoder-decoder framework for potential outcome estimation.}
    \label{fig:encoder_decoder}
\end{figure*}


\subsection{Pseudo-counterfactuals Generation}
Due to the unobserved nature of counterfactual samples, it brings difficult to train a model with good generalization performance. Especially in the process of time series forecasting task, only one decision is usually adopted at the present moment, while the results of other decisions are unknown. To explore the outcomes under other treatment options and the reasons for not taking them, it is necessary to complete the datasets at first.

Instead of learning a joint distribution of causal variables and treatments, we generate the pseudo-counterfactual data with the interventional features $\tilde{V}$ and treatment actions $\tilde{A}$. In detail, a corruption function $\mathbb{C}(\cdot)$ is proposed with $\mathbf{V}$ and $\mathbf{A}$ as input: 
\begin{small}
\begin{equation}
    (\tilde{\mathbf{V}},\tilde{\mathbf{A}}) = \mathbb{C}(\mathbf{V},\mathbf{A})\,.
\end{equation}
\end{small}

The corruption function is implemented based on experiment results, as described below. For each time step a per sample, we randomly specify an intervention position to be changed, then reverse operation is carried out on the intervention features and intervention actions on this position. The reverse operation can be like subtracting by identity vector, to obtain the corresponding counterfactual samples. Our experimental results prove that this strategy is effective.

\subsection{Individual Outcome Estimation}

The observational data can be used to train a supervised learning model to forecast the final outcome $\mathbf{Y}_{t}$. However, in the mixed intervention action scenario, the final result is produced by the combined action of multiple treatments, which is not conducive to discovering the role of each treatments/interventions. While potential outcomes under different treatment are unavailable, furthermore, due to the existence of time-varying confounders in time series, it introduces the time-varying bias which leads the model cannot be reliably used for making causal predictions. Hence, inspired by adversarial learning, TCFimt removes this bias through domain adversarial training and estimates the individual potential outcomes both on factual and counterfactual samples, for any intended future treatment assignment. 

The basic idea of individual outcome estimation is to learn a balanced treatment-invariant representation at each moment, and subsequently apply the balanced representation and the latest intervention information for potential outcome forecasting in future time steps. Here the balanced representation learning process is most critical. 

\textbf{Balanced representation.}
In time series data, we observed the historical information $\overleftarrow{\mathbf{H}}_{t} = (\overleftarrow{\mathbf{A}}_{t-1},
\overleftarrow{\mathbf{V}}_{t},
\overleftarrow{\mathbf{X}}_t)$ at time step $t$, which the historical time-varying variable $\overleftarrow{\mathbf{X}}_t$, intervention features $\overleftarrow{\mathbf{V}}_{t}$ and treatment assignment $\overleftarrow{\mathbf{A}}_{t-1}$ all have an impact on the next potential outcomes. Notice that the current treatment information is excluded from the balanced representation learning for better exploring the impact of the current intervention. To remove the temporal dependence of the treatments, we hope the balanced representation summarizes the past temporal states $\overleftarrow{\mathbf{H}}_{t}$, but is not predictive of the treatment $\mathbf{A}_t$. 

Let $\Phi$ denote the representation function that maps the history $\overleftarrow{\mathbf{H_t}}$ to a representation space $\mathcal{R}$. To obtain unbiased treatment effects, $\Phi$ needs to construct treatment invariant representations for each treatment $K$ such that $\mathbb{P}(\Phi(\overleftarrow{\mathbf{H}}_t) | A_k = 1) = \mathbb{P}(\Phi(\overleftarrow{\mathbf{H}}_t) | A_k = 0)$. To achieve this and to estimate the potential outcomes under a planned sequence of treatments, the domain adversarial training framework combing with a sequence-to-sequence architecture is proposed.  

\textbf{Encoder-decoder framework.}
This framework mainly contains two tasks, as shown in Figure~\ref{fig:encoder_decoder}.
By separately predicting the current treatment actions $\hat{\mathbf{A}}_t$ and the next individual outcomes under each treatment option, the balanced representation is learned in the hidden space via the recurrent networks.

Specifically, to estimate the future individual outcomes in the encoder module, we adopt the recurrent network (e.g., RNN) to capture the temporal states of observed factual samples as well as the generated pseudo-counterfactual ones. Intuitively, by inputting the treatment actions $\mathbf{A}(\mathbf{\tilde{A}})$, interventional features $\mathbf{V}(\mathbf{\tilde{V}})$ and time-varying variables $X$ into RNN module, the balanced representation is learned for subsequent prediction in two targets. 

To aid the representation, we define for time step $s\geq t$, $\overleftarrow{\mathbf{\Omega}}_{s} = (\overleftarrow{\mathbf{H}}_t,
\{\hat{\mathbf{A}}_j\}_{j=t}^{s-1}, \{\hat{Y}_{j}\}_{j=t+1}^s)$, where $ \hat{\mathbf{A}}_j$ is an estimated treatment assignment at time $j$, and $\{\hat{Y}_{j}\}_{j=t+1}^t=\emptyset$.

Let $\hat{Y}[\mathbf{a}_{t+\tau}]$ be the prediction of individual outcome under the mixed intervention on time step $t+\tau$. The function $\mathbb{G}_y$ makes outcome estimation based on the intervention from the last time step, whose parameter $\theta_{y}$ is shared on factual and counterfactual samples. 
\begin{small}
\begin{equation}
\begin{aligned}
    \mathbf{\hat{Y}}[\mathbf{a}_{t+\tau}] = \mathbb{G}_{y}(\Phi(\overleftarrow{\mathbf{\Omega}}_{t+\tau}),\mathbf{A}_{t+\tau}=\mathbf{a}_{t+\tau};\theta_{y})
\end{aligned}
\end{equation}
\end{small}

The final outcome is calculated by summarizing these estimated individual outcomes, then we can use the outcome to back-propagate.

In addition to the individual outcome estimation, another task is to classify the current treatment action, which forms a confrontation for learning the treatment invariant representation. Via $K$ classifiers $\mathbb{G}_{a}=(\mathbb{G}_{a_1},\mathbb{G}_{a_2},...,\mathbb{G}_{a_K})$ with parameter $\theta_{a}$, the treatment assigned at the current time is determined. 
\begin{small}
\begin{equation}
    \mathbf{\hat{A}}_{t} = \mathbb{G}_a(\Phi(\overleftarrow{\mathbf{{\Omega}}}_{t}); \theta_a)\,.
\end{equation}
\end{small}

Similarly, the decoder module also contains these two tasks, but it aims to predict the outcomes of unobserved samples for a sequence of future treatments, which is also called counterfactual prediction. To maintain the temporal information, the decoder network uses the balanced representation computed by the encoder to initialize the state of an RNN. Different from the encoder network, the decoder adopts the auto-regressive way to do predictions on account of the unavailable ground-truth outcomes. Specifically, it uses the (predicted) outcomes from the last time step like $\mathbf{\hat{Y}}_{t+1}$ and $\mathbf{\hat{Y}}_{t+2}$ combining with previously known treatment actions and intervention features as inputs. 

\subsection{Effect Disentanglement Block}
As the prediction from the above individual outcome estimation module is based on the balanced representation learned in a variety of intervention scenarios, it has limitations on the estimation of separated treatment effects and causal interactions. To tackle this challenge, we design an effect disentanglement block to rectify it. 

Thanks to the pseudo-counterfactuals generation module, we have factual samples and pseudo-counterfactual samples. Let $\hat{Y}^{F}[a_{k,t+1}]$ and $\hat{Y}^{CF}[\tilde{a}_{k,t+1}]$ be the factual and counterfactual individual outcome under the $k$-th intervention on time step $t+\tau$. The function $\mathbb{G}_y$ makes outcome estimation based on the intervention from the last time step, whose parameter $\theta_{y}$ is shared on factual and counterfactual samples. 
\begin{small}
\begin{equation}
\begin{aligned}
    \mathbf{\hat{Y}}^{F}[a_{k,t+1}] = \mathbb{G}_{y}(\Phi(\overleftarrow{\mathbf{H}}_{t+1}),\mathbf{A}_{t+1}=a_k;\theta_{y})\\
    \mathbf{\hat{Y}}^{CF}[\tilde{a}_{k,t+1}] = \mathbb{G}_{y}(\Phi(\overleftarrow{\mathbf{H}}_{t+1}), \mathbf{\tilde{A}}_{t+1}=\tilde{a}_k;\theta_{y})
\end{aligned}
\end{equation}
\end{small}

In order to disentangle the effect under multiple interventions and more accurately learn the relationship between the estimated outcomes and the corresponding interventions, we adopt a mutual information-based contrastive learning strategy.

In particular, as shown in the dashed circle of Figure~\ref{fig:training_process}, considering the learned balancing representation and the treatment information of factual data, we learn a medium representation and obtain an anchor to distinguish counterfactual from factual data in the estimated potential outcomes. The medium representation $\mathbf{Z}_t = (\mathbf{Z}_{1,t}, ..., \mathbf{Z}_{K,t})$ is calculated via a representation function $\Psi$ with dimension of the number of treatment options and a parameter $\theta_z$:

\begin{small}
\begin{equation}
    \mathbf{Z}_{k,t} = \Psi(\Phi(\overleftarrow{\mathbf{H}}_{t}), \mathbf{A}_{t}; \theta_z)\,.
\end{equation}
\end{small}

Then the anchor, denoted as $\mathbf{O}_t = (\mathbf{O}_{1,t}, ..., \mathbf{O}_{K,t})$, is calculated with the factual potential outcome $\mathbf{Y}^{F}$ as:
\begin{small}
\begin{equation}
    \mathbf{O}_{k,t} = \mathbf{A}_{k,t} \odot \mathbf{Z}_{k,t} \odot \mathbf{Y}^{F}[a_{k,t}]\,.
\end{equation}
\end{small}

Based on the anchor and inspired by CPC (contrastive predictive coding)\cite{oord2018representation}, to measure the relationship on the present state with specific intervention and next individual outcome, we calculate the mutual information $I(\mathbf{O}_{k,t}, \hat{\mathbf{Y}}[a_{k,t+1}])$ between them. For brevity, we denote $\mathbf{O}_{k,t}$ by $\mathbf{O}$ and $\hat{\mathbf{Y}}[a_{k,t+1}]$ by $\hat{\mathbf{Y}}$:
\begin{small}
\begin{equation}
    I(\mathbf{O}, \hat{\mathbf{Y}}) = \sum_{o,\hat{y}}p(\mathbf{O}, \hat{\mathbf{Y}})\log{\frac{p(\hat{\mathbf{Y}} \mid \mathbf{O})}{p(\hat{\mathbf{Y}})}}\,.
\end{equation}
\end{small}

Similar to the CPC method, we do not predict future potential outcome $\hat{Y}[a_{k,t+1}]$ directly with a generative model $p(\hat{Y}[a_{k,t+1}] \mid O_{k,t})$. Instead, we model a density ratio which preserves the mutual information between $\mathbf{O}_{k,t}$ and $\hat{\mathbf{Y}}[a_{k,t+1}]$ as:
\begin{small}
\begin{equation}
    f(\mathbf{O}_{k,t},\hat{\mathbf{Y}}[a_{k,t+1}]) \propto \frac{p(\hat{\mathbf{Y}}[a_{k,t+1}]\mid O_{k,t})}{p(\hat{\mathbf{Y}}[a_{k,t+1}])}\,.
\end{equation}
\end{small}

where $\propto$ means `proportional to'. The density ratio function is applied as an absolute difference model, to measure the distance between the anchor and individual treatment outcomes:
\begin{small}
\begin{equation}
     f(\mathbf{O}_{k,t},\hat{\mathbf{Y}}[a_{k,t+1}]) = \mid \mathbf{Z}_{k,t} \odot \hat{\mathbf{Y}}[a_{k,t+1}] - \mathbf{O}_{k,t} \mid\,.
\end{equation}
\end{small}

After estimating outcomes of multiple treatments $Y[\mathbf{a}_{t+1}]$ and outcomes of individual treatments $Y[a_{k,t+1}]$ , the causal interactions $\hat{\delta}_{CI}$ can be easily estimated for sample $i$ at time $t$ (by Assumption 2):
\begin{small}
\begin{equation}
\begin{aligned}
\hat{\delta}_{CI} &=(\hat{Y}[\mathbf{a}_{t+1}]^{(i)} - \hat{Y}[a_{0,t+1}]^{(i)}) \\
&- \sum_{k=0}^{K} (\hat{Y}[a_{k,t+1}]^{(i)} - \hat{Y}[a_{0,t+1}]^{(i)}) \,.
\end{aligned}
\end{equation}
\end{small}


\subsection{Training Procedure}
\begin{figure}
    \centering
    \includegraphics[scale=0.4]{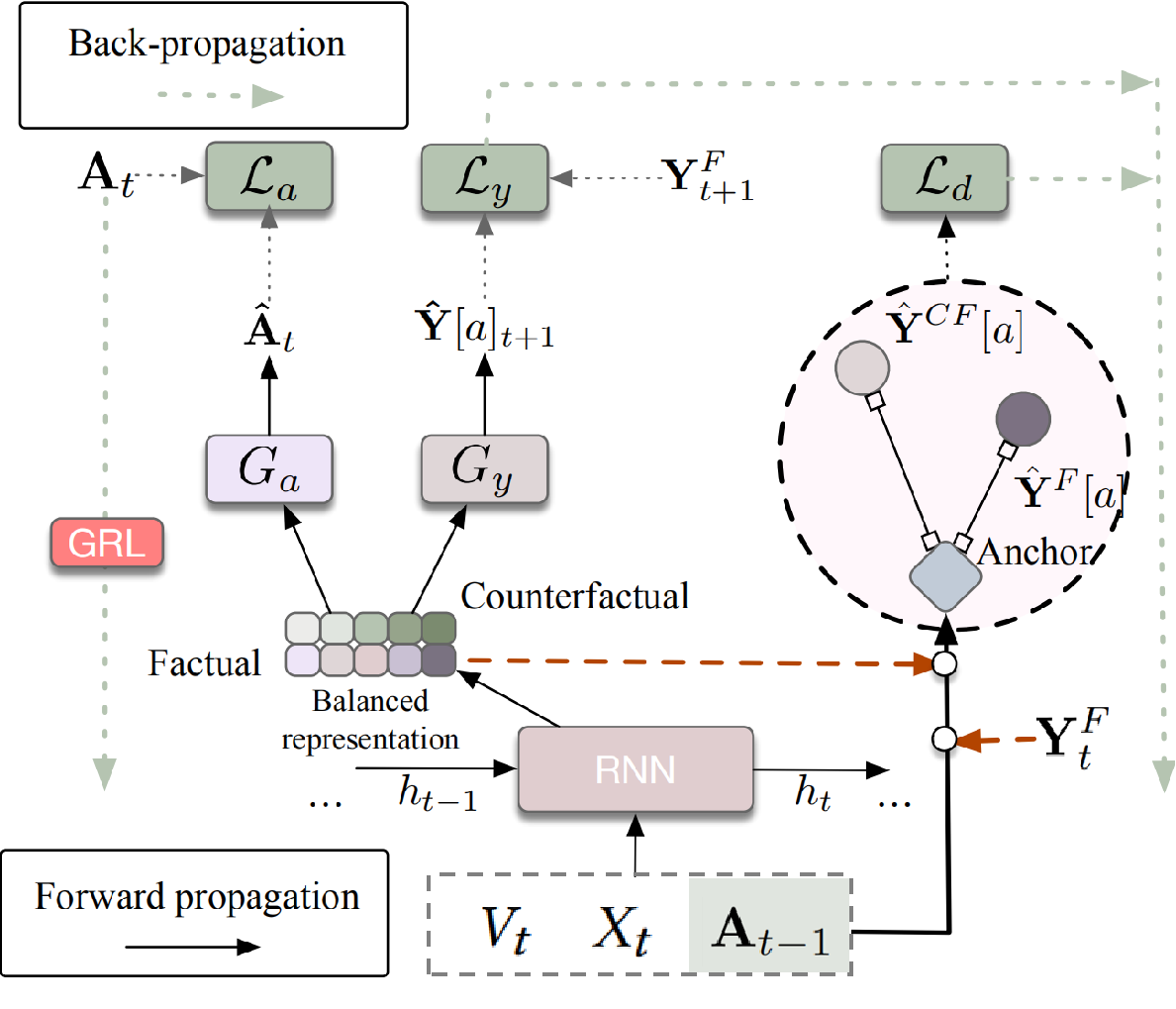}
    \caption{Training Procedure of the whole architecture.}
    \label{fig:training_process}
\end{figure}

In this subsection, we present the training details of our whole architecture. To remove the bias from time-dependent confounders, it requires that the distance in the distribution of $\Phi(\overleftarrow{\mathbf{H}}_{t})$ between any two pairs of treatments to be minimized. The normally used idea is to assume that the $K$ different treatments represent $K$ pairs of distinct domains ($K_{th}$ treatment implement or not), then based on an adversarial framework of domain adaption to build a representation, which achieves the maximum error in domain classification and the minimum error in outcome prediction. 

TCFimt uses domain adversarial training to build a representation of the history $\Phi(\overleftarrow{\mathbf{H}}_{t})$ that is both invariant to the treatment action $\mathbf{A}_t$ given at time step $t$. Specifically, the domain adversarial training is mainly realized through a GRL (Gradient Reversal Layer)~\cite{ganin2015unsupervised}, which leaves the input unchanged during forwarding propagation and reverses the gradient by multiplying it by a negative scalar during the back-propagation. Further, we add a self-supervised training loss for learning the disentangled intervention effects. The whole training procedure is shown in Figure~\ref{fig:training_process}. The total loss consists of three loss functions:


\begin{compactitem}
    \item The treatment classification loss $\mathcal{L}^{(i)}_{a,t}$ is as follows:
    \begin{small}
    \begin{equation}
        \mathcal{L}^{(i)}_{a,t} = - \sum_{k=1}^{K} \sum_{j=0,1} \mathbb{I}_{\{a_{k,t}^{(i)}=j\}}\log{(\mathbb{G}^{a_{k,t}=j}_a(\Phi(\overleftarrow{\mathbf{H}}_{t}; \theta_r); \theta_a))}\,.
    \end{equation}
    \end{small}
    \item The potential outcome forecasting loss $\mathcal{L}^{(i)}_{y,t}$ is as follows:
    \begin{small}
    \begin{equation}
        \mathcal{L}^{(i)}_{y,t} = || \mathbf{Y}^{(i)}_{t+1} -  \mathbb{G}_y(\Phi(\overleftarrow{\mathbf{H}}_{t}; \theta_r); \theta_y) ||^{2}\,.
    \end{equation}
    \end{small}
    
    \item The contrastive learning loss is based on InfoNCE loss:
    \begin{small}
    \begin{equation}
    \begin{aligned}
    \mathcal{L}^{(i)}_{d,t} &= - \sum_{k=0}^{K} \log\frac{f(\mathbf{O}_{k,t},\hat{\mathbf{Y}}^{F}[a_{k,t+1}])}{f(\mathbf{O}_{k,t},\hat{\mathbf{Y}}^{CF}[a_{k,t+1}])} \\
     & = - \sum_{k=0}^{K} \log \frac{\mid \mathbf{Z}_{k,t} \odot \hat{\mathbf{Y}}^{F}[a_{k,t+1}] - \mathbf{O}_{k,t} \mid}{\mid \mathbf{Z}_{k,t} \odot \hat{\mathbf{Y}}^{CF}[a_{k,t+1}] - \mathbf{O}_{k,t} \mid} \,.
    \end{aligned}
    \label{equ:effect_disen_loss}
    \end{equation}
    \end{small}
\end{compactitem}

In a nutshell, our final objective function is to maximize treatment loss, minimize potential outcome loss and minimize the InfoNCE loss~\cite{oord2018representation}. Thus, the overall loss $\mathcal{L}_{t}^{(i)}$ at timestep $t$ is given by:
\begin{small}
\begin{equation}
\begin{small}
    \mathcal{L}_{t}^{(i)}(\theta_r, \theta_a, \theta_y, \theta_z) = \sum_{i=1}^{N} \mathcal{L}^{(i)}_{y,t} - \lambda_1 \mathcal{L}^{(i)}_{a,t} + \lambda_2  \mathcal{L}^{(i)}_{d,t}\,.
    \label{equ:total_loss}
\end{small}
\end{equation}
\end{small}

The hyperparameters $\lambda_1$ and $\lambda_2$ are used to balance the loss scale.  For the choice of the hyperparameters, we start with an initial value for $\lambda_1$ and $\lambda_2$ use an exponentially increasing schedule during training and get relatively stable results.

\subsection{Theoretical Analysis}

In this section, we theoretically show that the developed method is effective. Specifically, we show that our overall loss function can effectively bound the expected error of multi-treatment effect estimation. To uncover this, we first show in Theorem 1 that the overall expected error of treatment effect estimation can be bounded by the sum of $\epsilon_F$ and $\epsilon_{CF}$, where $\epsilon_F$ is the expected factual loss and $\epsilon_{CF}$ is the expected counterfactual losses.  We note that $\epsilon_F$ is upper bounded by $\mathcal{L}^{(i)}_{y,t}$ and transform the inestimable $\epsilon_CF$ (because counterfactual data are not observed) to the difference, $\epsilon_{CF}$ - $\epsilon_F$, which is upper bounded by the discrepancy of treated and control distributions. We then show in Theorem 2 that the discrepancy can be upper bounded by $\mathcal{L}^{(i)}_{a,t}$. Therefore, the overall expected error of multi-treatment effect estimation can be bounded by the combination of the potential outcome forecasting loss $\mathcal{L}^{(i)}_{y,t}$ and the treatment classification loss $\mathcal{L}^{(i)}_{a,t}$ in the adversarial framework, which ensures the estimability of multiple treatment effects.
Finally, we show in Theorem 3 that minimizing the contrastive learning loss $\mathcal{L}^{(i)}_{d,t}$ maximizes the lower bound of mutual information of anchors and outcomes, which further enhances the balanced representation, facilitating the estimation of separated main treatment effects and causal interactions.

\textbf{Theorem 1. Upper Bound of Estimation} Let $\epsilon_{PEHE}(f)$ denote the expected error in estimating the individual treatment effect of a function\footnote{See the details of definition and proof in the supplementary material.}. Let $disc(.,.)$ denote the discrepancy, let $p^{a_{k,t}}_{\Phi}$ be treated and control distributions induced by $\Phi$ on $R$, then :
\begin{small}
\begin{equation}
\begin{aligned}
    &\epsilon_{PEHE}(h,\Phi)\\
    &\leq 2(\epsilon_{CF}(h,\Phi)+\epsilon_{F}(h,\Phi)-2\sigma^2_{Y})\\
    &\leq 2(\epsilon^{a_{k,t}=0}_{F}(h,\Phi)+\epsilon^{a_{k,t}=1}_{F}(h,\Phi)\\
    &+\mathop{disc}(p^{a_{k,t}=1}_{\Phi},p^{a_{k,t}=0}_{\Phi}))   
\end{aligned}
\end{equation}
\end{small}

According to Theorem 1, $\epsilon_{PEHE}$ is upper bounded by the sum of the expected factual loss $\epsilon_{F}$ and expected counterfactual loss $\epsilon_{CF}$. And discrepancy of treated and control distributions can be effective upper bound of the difference $\epsilon_{CF}-\epsilon_{F}$ of treatment $k$ at time $t$. The proof of Theorem 1 is based on the theory in CFR~\cite{shalit2017estimating}. 

\textbf{Theorem 2. Balanced representation} let $P_j$ denote the distribution of $H_t$ conditional on $a_{k,t} = i$, Let
$\mathbb{G}^{a_k=i}_a$ denote the output of $\mathbb{G}_a$ corresponding to treatment $a_{k,t} = i$. Then the objective for $\Phi$ becomes:

\begin{small}
\begin{equation}
\begin{aligned}
    \mathop{min}\limits_{\Phi}
    \mathop{max}\limits_{G_{a}}
    \sum_{k=1}^{K}
    \sum_{i=0,1}
    \mathbb{E}_{\overleftarrow{\mathbf{H}}_{t}\sim P_{a_k}} 
 [\log{(\mathbb{G}^{a_k=i}_a(\Phi(\overleftarrow{\mathbf{H}}_{t}; \theta_r); \theta_a))}]\\
    s.t. \quad k \in {1,2,..,K},\sum_{i=0,1}\mathbb{G}^{a_k=i}_a(\Phi(\overleftarrow{\mathbf{H}}_{t}))=1 \\
    =\sum^K_{k=1}JSD(P^{a_k=0}_{\Phi}(x')\Vert P^{a_k=1}_{\Phi}(x'))-2Klog2.\\
\end{aligned}
\end{equation}
\end{small}

The result of Theorem 2 shows $\mathcal{L}^{(i)}_{a,t}$ has a global minimum which is attained if and only $P_{\Phi}^{a_k=0}=P_{\Phi}^{a_k=1}$ for each $k \in {1,2,..,K}$. And $\mathcal{L}^{(i)}_{a,t}$ can be replacement of Jensen-Shannon Divergence $JSD(\cdot \Vert \cdot)$ without the time dependency. Combined with the result of Theorem 1 and Theorem 2, $\mathcal{L}^{(i)}_{a,t}$ can be the upper bound of the difference $\epsilon_{CF}-\epsilon_{F}$ in temporal multi-intervention. 

\textbf{Theorem 3. Estimating the Mutual Information with InfoNCE} Let $I(\cdot , \cdot)$ denote mutual information, then the infoNCE loss:
\begin{small}
\begin{equation}
\begin{aligned}
  \mathcal{L}^{(i)}_{d,t}
  &= - \sum_{k=0}^{K} \log\frac{f(\mathbf{O}_{k,t},\hat{\mathbf{Y}}^{F}[a_{k,t+1}])}{f(\mathbf{O}_{k,t},\hat{\mathbf{Y}}^{CF}[a_{k,t+1}])} \\
  &\geq -\sum_{k=0}^{K}I(\mathbf{O}_{k,t},\hat{\mathbf{Y}}^{F}[a_{k,t+1}]).
\end{aligned}
\end{equation}
\end{small}

Theorem 3 shows the infoNCE loss is a low bound of mutual information between $\mathbf{O}_{k,t}$ and $\hat{\mathbf{Y}}^{F}[a_{k,t+1}]$. The result proves that $\mathcal{L}^{(i)}_{d,t}$ enhances the representation learning in the estimation of separated treatment outcomes and causal interactions. For details, see the supplementary material.

\section{Experiments}
We mainly focus on the following research questions:

\begin{compactitem}
    \item[Q1:] How is the prediction performance of the proposed TCFimt method for future outcomes?
    \item[Q2:] What is the auxiliary ability of the TCFimt method for future intervention decision making and the suitable timing of intervention?
    \item[Q3:] After the separation of intervention effects, what is the prediction performance of causal interactions between different treatments?
\end{compactitem}

\subsection{Experimental Settings}

\textbf{Datasets}: For the sake of fully verifying the above questions, we conduct experiments on real-world datasets from two distinct fields. 

\textit{Games}: The dataset contains time-series gaming indicators in distinct game genres coming from NetEase Games. For each game, there is a corresponding game update announcement data, which can be considered as interventions. According to the type of events in the game update announcement, we can get the intervention actions or event options, as summarized in Table~\ref{tab:game_statistics}. These two datasets are all collected from Jan. 2019 to Nov. 2020.

\textit{Tumour}: With the help of a state-of-the-art bio-mathematical model, this dataset simulates the combined effects of chemotherapy and radiotherapy in lung cancer patients. And at each timestep, there are four treatment options: no treatment, chemotherapy, radiotherapy, combined chemotherapy and radiotherapy. To validate the effectiveness, we adopt the model was based on related wrok~\cite{lim2018forecasting}~\cite{bica2020estimating} to evaluate model\footnote{The details of Tumour model in the supplementary material}.

\begin{table}
    \centering
    \resizebox{.47\textwidth}{!}{
    \begin{tabular}{c c c c}
         \hline
         Game ID & \#Events & \#Event Options & Game Genres \\ 
         \hline\hline
         Game1 & 695 & 3  & action game\\ 
         Game2 & 691 & 2 & role-playing game\\
        \hline
    \end{tabular}
    }
    \caption{The Statistics of Games Dataset}
    \label{tab:game_statistics}
\end{table}


\textbf{Baselines}:
For a fair comparison, we pick baseline methods from three categories. The first group includes classical and widely used time series prediction methods, i.e., ARMA, ARIMA, and Prophet. The second one is the deep neural network related sequential forecasting methods, i.e., LSTM and TCN. From the individual treatment effect estimation perspective, the last type of methods are the latest research approach in this area, like TARNet, CFR, and CRN.
\begin{compactitem}
    \item \textbf{ARMA}~\cite{mcleod1975derivation} is the autoregressive moving average method.
    \item \textbf{ARIMA}~\cite{zhang2003time} is the autoregressive integrated moving average method.
    \item \textbf{Prophet}~\cite{taylor2018forecasting} was proposed by Facebook, is based on decomposition way. 
    \item \textbf{LSTM}~\cite{hochreiter1997long} is a basic recurrent neural network method for sequential prediction.
    \item \textbf{TCN}~\cite{bai2018empirical} combines the convolutional and recurrent networks for sequential modeling.
    \item \textbf{TARNet\footnote{https://github.com/clinicalml/cfrnet}}~\cite{johansson2016learning} is the treatment-agnostic representation network, to learn a representation that reduce the discrepancy between the treated and control populations.
    \item \textbf{CFR}\footnote{https://github.com/clinicalml/cfrnet}~\cite{shalit2017estimating} is the counterfactual regression network, which is similar to TARNet but adding an integral probability metric for bounding the couterfactual loss.
    \item \textbf{CRN}\footnote{https://github.com/ioanabica/Counterfactual-Recurrent-Network/}~\cite{bica2020estimating} is the latest counterfactual regression network.
\end{compactitem}

\textbf{Evaluation Metrics}: For evaluating the forecasting performance in time series, we adopt the normally used RMSE and MAE metrics. As the counterfactual is not available in the games dataset, we treat the testing set in the future time steps as counterfactual samples. Considering question Q2 and Q3, the external accuracy of recommending specific treatment and treatment timing is adopted. 

\textbf{Parameter Settings}: Our TCFimt model is implemented with Tensorflow and optimizes the final objective function (in Eq.~\ref{equ:total_loss}) via the Adam algorithm. The hyperparameters like batch size, learning rate, hidden units in the RNN module, the dropout probability, etc. are searching from a range of values. Details of our implementations can be found in the supplementary material.

\subsection{Experimental Results}

\textbf{Counterfactual Forecasting Performance.} To demonstrate the effectiveness of the TCFimt method on future forecasting for counterfactual samples, we compare it with the above baselines on the Games dataset with two percentage metrics, as shown in Table~\ref{tab:factual_performance}. Since the real intervention are obtained with time-series data in this dataset, the comparison result is more authentic and reliable. From the table, we have the following observations:

The TCFimt achieves the best performance on the Game1 dataset. For the Game2 dataset, our proposed method also beats all baselines on the RMSE metric and is slightly inferior to the first type of comparison method on the MAE metric. This illustrates that the learned balanced representation by TCFimt is more effective in predicting the potential outcomes of subsequent counterfactual samples.

Through the data analysis of Game1 and Game2 datasets, the fluctuations in the Game2 dataset are greater than that of Game1. Therefore, from the comparison of the MAE results on Game2, the ARMA and ARIMA model are more inclined to predict an average result, which leads to the corresponding MAE result is smaller.

\begin{table}
    \centering
    \resizebox{.47\textwidth}{!}{
    \begin{tabular}{c|c|c|c|c}
          \toprule
          \hline
        \multirow{2}{*}{Method} &\multicolumn{2}{c|}{\textbf{Game1}} & \multicolumn{2}{c}{\textbf{Game2}}\\
        \cline{2-5}
        & RMSE(\%) & MAE(\%) & RMSE(\%) & MAE(\%)\\
        \hline\hline
        ARMA & 6.75 & 3.88 & 7.11 & 3.72 \\
        ARIMA & 6.52 & 4.13 & 6.96 & 3.92 \\
        Prophet & 5.91 & 3.84 & 6.29 & 4.11 \\
        \hline
        LSTM & 6.20 & 3.62 & 7.08 & 4.36 \\
        TCN & 5.93 & 4.69 & 6.43 & 5.04 \\
        \hline
        CFR & 6.69 & 4.75 & 8.04 & 4.90 \\
        TARNet & 6.73 & 4.92 & 8.02 & 4.98 \\
        CRN & 5.94 & 4.85 & 6.05 & 4.31\\
        TCFimt(Ours) & \textbf{4.78}& \textbf{3.60} & \textbf{5.95} &  \textbf{4.09}\\
        \hline
        \bottomrule
    \end{tabular}}
    \caption{Performance of next-one factual prediction on two game datasets.}
    \label{tab:factual_performance}
\end{table}

\textbf{Treatment Recommending Performance.} To validate the second question Q2 and Q3, we conduct experiments on the Tumour dataset as to which treatment actions were taken and when\footnote{See the supplementary material for more details.}. We first calculate the RMSE of the prediction of the tumour volume and causal interactions by assumption 2. We then calculate the accuracy of taking the right treatments (Tr Acc.) for patients, and then the accuracy of treatment timing (TrT Acc.) when the right treatment was selected. Based on the such rule, the TrT Acc. is bound to be smaller than Tr Acc. In Table~\ref{tab:acc_performance}, the parameter $\tau$ means the predicted horizon, $\gamma_c$ and $\gamma_r$ are the hyperparameters to generate distinct Tumour datasets with varied treatment assignments. 
The results show that our proposed method TCFimt achieves higher accuracy than the CRN method in most cases, i.e., the TCFimt model has certain advantages in assisting the formulation of treatment. 

\begin{table}
    \centering
    \resizebox{.47\textwidth}{!}{ 
    \begin{tabular}{c|c|c c|c c|c c}
          \toprule
        \hline
        \multicolumn{2}{c|}{} & \multicolumn{2}{c|}{$\gamma_c = 5, \gamma_r = 5$} & \multicolumn{2}{c|}{$\gamma_c = 5, \gamma_r = 0$} & \multicolumn{2}{c}{$\gamma_c = 0, \gamma_r = 5$} \\
        
        \hline
        & $\tau$ & CRN & TCFimt & CRN & TCFimt & CRN & TCFimt\\
        \hline \hline
       \multirow{5}{*}{RMSE} & 3 & 6.65\% & \textbf{5.92}\% & 2.07\% & \textbf{2.02\%} & 3.09\% & \textbf{2.44}\% \\
        & 4 & 7.94\% & \textbf{6.23\%}  & 2.08\% & \textbf{1.96\%} & 2.46\% & \textbf{2.32}\% \\
        & 5 & 8.03\% & \textbf{7.26\%} & 2.41\% & \textbf{2.26\%} & 3.06\% & \textbf{2.66\%}  \\
        & 6 & 7.23\% & \textbf{7.02\%} & 2.67\% & \textbf{2.42\%} & 2.76\% & \textbf{2.48\%}   \\
        & 7 & 9.62\% & \textbf{8.14\%} & 2.83\% & \textbf{2.58\%} & 2.81\% & \textbf{2.74\%} \\
        \hline
       \multirow{5}{*}{Tr Acc.} & 3 & 63.0\% & \textbf{74.6\%} & 69.4\% & \textbf{73.4\%} & 69.6\% & \textbf{71.4\%} \\
        & 4 & 65.2\% & \textbf{73.2\%} & 71.2\% & \textbf{72.9\%} & 70.8\% & \textbf{71.0}\% \\
        & 5 & 65.4\% & \textbf{72.4\%} & 70.0\% & \textbf{71.8\%} & 65.2\% & \textbf{69.4\%}  \\
        & 6 & 66.2\% & \textbf{71.0\%} & 65.8\% & \textbf{70.2\%} & 65.8\% & \textbf{69.3\%}   \\
        & 7 & 63.4\% & \textbf{70.2\%} & 65.2\% & \textbf{70.2\%} & 66.2\% & \textbf{68.8\%} \\
        \hline
       \multirow{5}{*}{TrT Acc.} & 3 & 36.0\% & \textbf{38.8\%} & 29.6\% & \textbf{38.8\%} & 35.0\% & \textbf{35.8\%} \\
        & 4 & 34.0\% & \textbf{36.0\%} & 35.9\% & \textbf{36.6\%} & 34.0\% & \textbf{36.8\%} \\
        & 5 & 35.6\% & \textbf{35.8\%} & 35.8\% & \textbf{37.6\%} & 31.2\% & \textbf{36.4\%} \\
        & 6 & 36.0\% & \textbf{36.8\%} & 34.0\% & \textbf{36.0\%} & 33.8\% & \textbf{35.3\%} \\
        & 7 & 35.2\% & \textbf{37.2\%} & 35.2\% & \textbf{36.8\%} & 39.8\% & \textbf{39.9\%} \\
        \bottomrule
        \hline
    \end{tabular}
    }
    \caption{Performance of recommending the correct treatment and timing of treatment on Tumour dataset.}
    \label{tab:acc_performance}
\end{table}

\section{Conclusion}
In this paper, we propose a novel method named TCFimt for counterfactual forecasting of time-series data in multiple mixed intervention scenarios. We first design a function to generate pseudo counterfactual samples, which helps to mitigate the selection bias. Via the adversarial loss of potential outcome prediction and treatment classification tasks, a treatment invariant representation was learned for alleviating the time-varying bias. Finally, the intervention mixing problem is solved based on a contrastive learning way, which also enhanced the forecasting of individual potential outcomes. Extensive experiments exhibited the superiority of our proposed model on counterfactual forecasting and choosing the correct treatments. Our study is an attempt at causal inference under mixed treatment effects, it may bring some new insights in this direction.  






\bibliography{aaai23}

\clearpage

\end{document}